\documentclass[10pt,twocolumn,letterpaper]{article}

\usepackage{iccv}
\usepackage{times}
\usepackage{epsfig}
\usepackage{graphicx}
\usepackage{amsmath}
\usepackage{amssymb}

\usepackage[pagebackref=true,breaklinks=true,letterpaper=true,colorlinks,bookmarks=false]{hyperref}

\usepackage{bbm}
\usepackage{multirow}
\usepackage[table,xcdraw]{xcolor}
\usepackage{textcomp}
\usepackage{enumitem}
\usepackage{pifont}

\iccvfinalcopy 

\def\iccvPaperID{3052} 

\ificcvfinal\fi

\begin{document}

\title{CoReFace: Sample-Guided Contrastive Regularization for Deep Face Recognition}

\author{Youzhe Song\\
East China Normal University\\
Shanghai, China\\
{\tt\small yanfengz@outlook.com}
\and
Feng Wang\\
East China Normal University\\
Shanghai, China\\
{\tt\small fwang@cs.ecnu.edu.cn}
}

\maketitle
\ificcvfinal\thispagestyle{empty}\fi

\begin{abstract}
The discriminability of feature representation is the key to open-set face recognition.
Previous methods rely on the learnable weights of the classification layer that represent the identities.
However, the evaluation process learns no identity representation and drops the classifier from training.
This inconsistency could confuse the feature encoder in understanding the evaluation goal and hinder the effect of identity-based methods.
To alleviate the above problem, we propose a novel approach namely Contrastive Regularization for Face recognition (CoReFace) to apply image-level regularization in feature representation learning.
Specifically, we employ sample-guided contrastive learning to regularize the training with the image-image relationship directly, which is consistent with the evaluation process.
To integrate contrastive learning into face recognition, we augment embeddings instead of images to avoid the image quality degradation. 
Then, we propose a novel contrastive loss for the representation distribution
by incorporating an adaptive margin and a supervised contrastive mask to generate steady loss values and avoid the collision with the classification supervision signal.
Finally, we discover and solve the semantically repetitive signal problem in contrastive learning by exploring new pair coupling protocols.
Extensive experiments demonstrate the efficacy and efficiency of our CoReFace which is highly competitive with the state-of-the-art approaches.
\end{abstract}

\vspace{-.2in}
\section{Introduction}
Face recognition (FR) is a long-standing task and plays an important role in numerous applications. The evaluation scenarios of FR could be categorized into two types, i.e. verification and identification. Both of them are based on the similarity between face images.
To better adapt to the realistic situations, the identities for evaluation are excluded from the training in open-set face recognition~\cite{huang_lfw_2007}.
Recently classification methods achieve the state-of-the-art (SoTA) results in FR where the face images with identity labels are used to train a fine-grained classifier to discriminate different identities. While in evaluation, the classifier is usually dropped as the training-specific identity information contributes little to this process.


\begin{figure}
    \centering
    \includegraphics[scale=0.5]{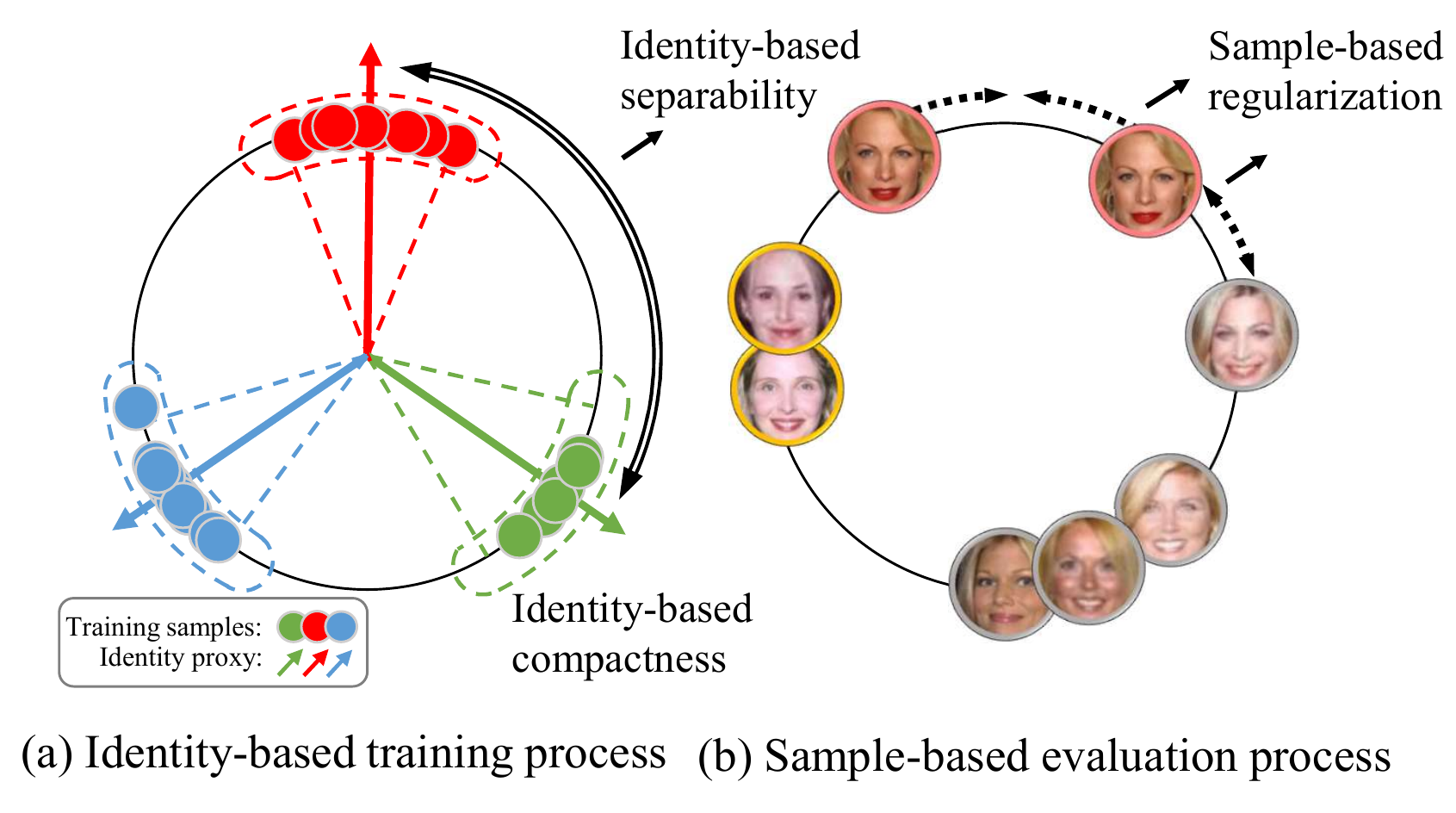}
    
    \caption{
    (a) Current identity-based methods aim at intra-class compactness and inter-class separability in training. (b) However, the identity-centric training pays little attention to image-image relationship which is the foundation in evaluation process. The points with grey borderlines are from distinct identities, and other points with the same borderline color are from the same identity. Our CoReFace takes contrastive learning as regularization to directly constrain the relationship between images during training.
    }
    \label{fig:theory}
    \vspace{-.2in}

\end{figure}

To achieve higher intra-class compactness, a series of margin-based methods are proposed~\cite{liu_sphereface_2017, wang_cosface_2018,deng_arcface_2019}, which put a margin to make the decision of the right class harder in training.
However, these classification methods ignore the holistic feature space~\cite{uniformface}.
Some other works focus more on the inter-class separability which is also a key for the feature discriminability, and design different loss functions to perform regularization~\cite{center_loss,regularface,yang_orthogonality_2021,uniformface}.
However, they only investigate the image-identity or the identity-identity relationships during training, and do not fully constrain the image-image similarity which is essential in evaluation. 
As illustrated in Figure~\ref{fig:theory}, the feature distribution in the identity-based training could achieve high intra-class compactness and inter-class separability with the help of identity proxy features.
However, in the sample-based evaluation, the classifier is dropped. Furthermore, the face images in evaluation are of the identities which are different with the training. Thus, the feature distribution in evaluation might be not as discriminative as in training.

To address the above problem, in this paper, we propose a novel approach namely Contrastive Regularization for Face recognition (CoReFace). We constrain the image-image relationship by using contrastive learning to regularize the training process so as to make the goal of training consistently with the evaluation, and thus boost the performance of open-set face recognition.
Contrastive learning pulls the semantically similar samples closer and pushes the others away in the representation space~\cite{contrastive_loss}.
In the FR literature, the class-guided contrastive learning has been attempted which takes samples from the \emph{same class} to compose positive pairs. For instance, triplet loss is applied solely~\cite{schroff_facenet_2015} or jointly~\cite{deepid2,deepid2+} with the classification methods.
However, with the recent development of margin-based methods, these approaches might cause interference in joint training with other classification methods~\cite{softmax+,deng_arcface_2019}.
On the other hand, sample-guided contrastive learning demonstrates a promising advancement in unsupervised learning~\cite{invaspread, insta_disc, chen_simple_2020}. They apply stochastic data augmentation on the \emph{same image} to compose positive pairs, which alleviates the limitation of the label requirement. It further provides a perspective that beyond class boundary.
In our approach, we employ sample-guided contrastive learning as regularization to adjust the image-image relationship for more semantic and consistent feature distribution in training and evaluation.


However, it is non-trivial to integrate the sample-guided contrastive learning with the margin-based classification methods.
First,
as a fine-grained task, face recognition requires a huge number of high-quality images to learn the difference between identities. The commonly-used data augmentations in contrastive learning hinder the convergence of FR models~\cite{urf, kim_adaface_2022}.
To make the sample-guided contrastive learning applicable to FR, we propose a new pipeline by using feature augmentation instead of data augmentation to generate positive pairs.
Second,
the sample-guided contrastive learning is usually designed to be solely applied. When jointly training with margin-based classification methods, we find their effectiveness become insignificant. To solve this problem, we design a novel contrastive loss function to effectively perform the regularization.
Third, the scale of the negative sample pool plays a key role in contrastive learning~\cite{chen_simple_2020,moco,simsiamese,gao_simcse_2021}. When we focus on a general situation with normal batch size and no extra encoder, a \emph{Semantically Repetitive Signal} (SRS) problem is discovered, i.e. some sample combinations repeatedly contribute to the optimization. This pushes the relative part of distribution with inappropriate magnitude. 
To alleviate this problem, we explore new strategies of pair coupling.
The main contributions of this paper are summarized as follows:
\begin{itemize}
\vspace{-.1in}
    \item We propose a novel framework to apply regularization in FR by contrastive learning.
    Unlike previous regularization approaches which adjust the feature distribution with image-identity pairs, our method utilizes image-image relationships which is consistent between training and evaluation.
   \vspace{-.1in}
    \item We propose a contrastive loss function to perform effective regularization which incorporates an adaptive margin to strengthen the contrastive supervision signal, and a supervised contrastive mask to avoid the supervision collisions in joint training.
    \vspace{-.1in}
    \item We investigate the SRS problem in contrastive learning in the situation of limited negative samples, and explore different pair coupling protocols to alleviate this problem.
    \vspace{-.1in}
    \item We conduct extensive experiments on the widely-used benchmarks to demonstrate the superiority of our proposed framework over the existing approaches.
\end{itemize}


\section{Related Works}
\label{related_works}

\subsection{Margin-based Classification methods}

In recent years, we have witnessed an arising trend on margin-based classification methods in FR~\cite{liu_sphereface_2017,wang_cosface_2018,deng_arcface_2019,wang_mis-classified_2020,huang_curricularface_2020}.
Among them, the representation embeddings of the images and the classes are normalized before their multiplication~\cite{normface, l2}, and then the product degrades to the cosine value of the angle between the two vectors.
During training, a margin parameter is taken to enlarge the distance between the matched image-identity pair and the unrelated ones. This improves the compactness of the intra-class with shorter distance between the representations of the same identity.
The normalization relieves the misguidance of the feature norm in the Softmax loss by projecting the features onto a hypersphere, and the margin puts a strong constraint on the image-identity feature pairs on this hypersphere. While these methods achieve high intra-class compactness, they fail to exploit the holistic feature space~\cite{uniformface}. To improve the feature distribution, our CoReFace puts constraints on the image-image relationship during training.

\begin{figure*}
    \centering
    \includegraphics[scale=0.65]{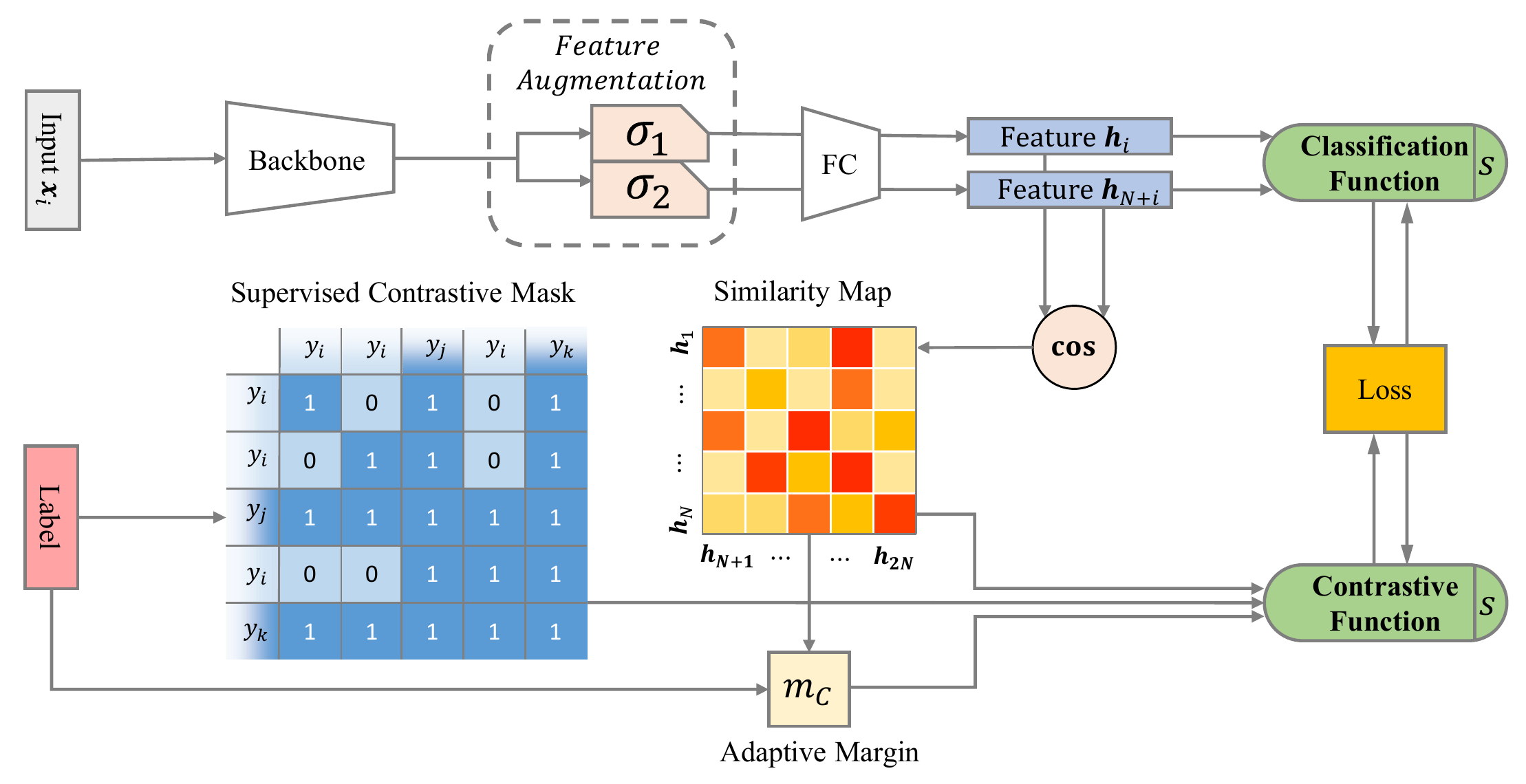}
     \vspace{-.05in}
    \caption{Illustration of our CoReFace approach.
    To relieve the image quality degradation problem, we add a feature augmentation module between the backbone and the FC layer to generate positive pairs for sample-based contrastive learning.
    Our contrastive loss function is composed by an adaptive-margin-based loss and a supervised contrastive mask. The margin is adaptive with the training process and the backbone magnitude. The supervised contrastive mask avoids the conflict between samples from the same identity in contrastive learning.
    We also take a new pair coupling protocol in the similarity computation for contrastive learning to avoid the semantically repetitive signal problem. The contrastive loss is then used to regularize the training process with the image-image relationship.
    }
    \label{fig:contra-framework}
    \vspace{-.15in}
\end{figure*}

\subsection{Feature Regularization in FR}


Feature distribution is the foundation of face recognition evaluation since both the two sub-tasks (verification and identification) rely on feature similarity between face images~\cite{liu_sphereface_2017,center_loss}.
To adjust the feature distribution in a holistic view, some methods resort to extra constraints to promote the performance of evaluation.
They restrict the magnitudes of the representation features~\cite{ring_loss}, or the Euclidean distance between the representations and the identity weights~\cite{center_loss}.
As the identity weights serve as the class proxies, a number of works argue that they could support the holistic feature distribution~\cite{range_loss,regularface,uniformface,yang_orthogonality_2021}.
By constraining the energy function, the Euclidean distances, or the angulars between identity weights, better distribution could be achieved.

Nevertheless, all of the above methods utilize the training-specific identity information to adjust the sample similarities indirectly. Their efficacy is designed for training with little assurance to generalize to the evaluation process where the identities are dropped.
In this paper, we propose a novel constrastive regularization approach by designing a contrastive loss.
Compared with existing approaches, we directly adjust the relationship of image features so as to make the training consistent with the evaluation, and thus improve the performance of FR under open-set situations.



\subsection{Contrastive Learning for FR}

Contrastive learning aims at clustering semantic neighbors as distribution neighbors in the representation space~\cite{contrastive_loss}.
Class-guided contrastive learning~\cite{schroff_facenet_2015, normface} has been applied to FR~\cite{schroff_facenet_2015}, which takes the samples from the same class as semantic neighbors. However, they have shown to obstruct the performance in joint training~\cite{deng_arcface_2019,softmax+}.
On the other hand, sample-guided contrastive methods take the outcome of data augmentation to compose positive pairs. They usually construct a large negative sample pool for comparison~\cite{chen_simple_2020,moco,simsiamese,gao_simcse_2021}. With huge dataset and sufficient training, they show promising performance on unsupervised learning.
However, it is hard to apply sample-guided constrastive learning in FR which would be trapped in the obstacles introduced by the commonly-used data augmentation~\cite{urf}.

In this paper, we design a new framework to reconcile the image quality degradation problem and keep regularization effective in the training stage.
CoReFace takes feature augmentation to avoid the semantic damage of data augmentation. In addition, the proposed contrastive loss adopts an adaptive margin to supervise the well-performed classification methods, and adopts a supervised contrastive mask to prevent the conflict in joint training.
We further discover the SRS problem in common FR training settings and explore pair-coupling protocols to relieve this problem.

\section{Methodology}

Figure~\ref{fig:contra-framework} illustrates the framework of our CoReFace.
We apply regularization with sample-guided contrastive learning to solve the neglect of image-image relationship in training and the inconsistency caused by the abandoning of the classifier in evaluation.
First, to address the image quality degradation problem, we employ feature augmentation to replace the widely-used data augmentation for positive pair composition.
We also drop the projection layer which is widely used with contrastive learning~\cite{chen_simple_2020, mocov2, byol, simsiamese}.
In our scenario, contrastive learning aims at adjusting the feature representation distribution, instead of an information-limited projection.
Second, we propose a novel contrastive loss by integrating an adaptive margin and a supervised contrastive mask. The adaptive margin is designed to keep the magnitudes of the positive and the negative similarities close, and produce steady loss values during the joint training. The supervised contrastive mask (SCM) takes the class label to generate a mask which excludes the samples of the same class from the negative comparison pool. This avoids the conflict with the classification method.
Third, we investigate the \emph{Semantically Repetitive Signal} (SRS) problem, i.e. some key pairs in contrastive learning are repeated. This distorts the feature distribution and disturbs the upcomming similarity calculation. We design new pair-coupling strategies to relieve this problem.
Finally, we apply image-image regularization to FR by jointly training the classification method with our CoReFace loss function.

\subsection{Feature Augmentation}
Data augmentations such as cropping with resizing, color distortion, cutout, and Gaussian blur are widely used to generate positive pairs in sample-guided contrastive learning in computer vision tasks. This would inevitably bring semantic damages to the samples and degrade the image qualities.
It is applicable to take strong augmentation in the coarse-grained classification tasks since the difference between images is relatively large.
However, FR is a fine-grained task and requires the face images to be semantically clear. The image quality degradation caused by data augmentation is not negligible~\cite{urf,kim_adaface_2022}.

To solve the above problem and make the sample-guided contrastive learning applicable in FR, we augment the features (instead of the images) for positive pair composition.
As illustrated in Figure~\ref{fig:contra-framework}, we pass the hidden embedding after the backbone through two dropout channels $\sigma_{1}$ and $\sigma_{2}$ with distinct masks.
Dropout~\cite{dropout} randomly mutes some part of the input with a certain probability. It can be seen as a kind of augmentation between two adjacent layers~\cite{dropaug, gao_simcse_2021}.
When dropout is applied on the input image, it could be thought as an extreme case of salt-and-pepper noise.
In our approach, the dropout masks are randomly generated in every mini-batch and operate on all of the input samples.
Other methods such as random noise~\cite{simgcl} is also suitable in our framework.

With feature augmentation, we can compose the positive pair for contrastive learning while avoiding the image quality degradation problem.
In addition, compared with data augmentation which is performed on the input sample, and passes the augmented samples to the whole model twice, our feature augmentation operates on the feature and saves nearly a half computation.

\subsection{CoReFace Loss Function} \label{CoReFace}

In our framework, contrastive loss is used to constrain the distribution of the features by providing the image-level distribution guidance to compensate the inconsistency between the identity-based training and the sample-based evaluation.
However, we find that the prevalent contrastive methods fail to keep effective signals in experiment. They insistently produce zero loss values and contribute little to the training. This is probably because that the classification method dominates the training by taking the advantage of labels. Furthermore, an aggressive regularization which conflicts with other supervision signals cannot work appropriately either.
To meet the above requirements, we design a novel contrastive loss function which is adaptively effective and harmonic in joint training. Our CoReFace loss can generate steady loss values and take the classification labels into consideration to avoid the collisions with the classification loss.

Both the sample-guided contrastive loss functions and the classification loss functions in FR are based on the cross-entropy loss function. The common forms of these two kinds of losses are as follows:
\begin{small}
    \begin{align}
    \mathcal{L}_{Cla} &= - \log 
    \frac{e^{s\cdot P(\boldsymbol{h}_i, \boldsymbol{W}_{y_i})}}
        {e^{s\cdot P(\boldsymbol{h}_i, \boldsymbol{W}_{y_i})} + \sum^{n}_{j=1, j \neq y_i}e^{s\cdot Q(\boldsymbol{h}_i, \boldsymbol{W}_j)}},
        \label{eq:softmax}\\
    \mathcal{L}_{Con} &= - \log 
    \frac{e^{\mathrm{sim}(\boldsymbol{h}_i, \boldsymbol{h}_{N+i})/\tau}}
        {\sum^{2N}_{j=1}\mathbbm{1}_{[j\neq i]}e^{\mathrm{sim}(\boldsymbol{h}_i, \boldsymbol{h}_j)/\tau}},
    \label{eq:contrastive}
    \end{align}
\end{small}
where $P(\boldsymbol{h}_i, \boldsymbol{W}_{y_i})$ and $Q(\boldsymbol{h}_i, \boldsymbol{W}_j)$ are two different functions to modulate the positive and the negative pair production of the feature $\boldsymbol{h} \in \mathbb{R}^d$,
$\boldsymbol{W} \in \mathbb{R}^{d\times n}$ is the weight of the classifier with $d$ being the feature dimension and $n$ being the number of classes,
$\mathrm{sim}(\boldsymbol{h}_i, \boldsymbol{h}_j)=\frac{\boldsymbol{h}^\top_i \boldsymbol{h}_j}{\Vert \boldsymbol{h}_i\Vert \Vert \boldsymbol{h}_j\Vert}$ is the cosine similarity,
$s$ and $\tau$ are two scale parameters used in the classification loss function and the contrastive loss function respectively, and
$\mathbbm{1}_{[j\neq i]} \in \{0,1\}$ is an indicator function evaluating to 1 iff $j \neq i$.


\textbf{Adaptive Margin.}
We follow the margin-based methods~\cite{liu_sphereface_2017,wang_cosface_2018,deng_arcface_2019} to enlarge the similarity of the positive pair and the dissimilarity of the negative pairs by increasing the difficulty of the judgement with a margin parameter $m$. Different from the classification methods which take the image-class pairs, our contrastive method takes image-image pairs. In this way, we alleviate the inconsistency between training and evaluation discussed above. Furthermore, we dynamically adjust the margin parameter during training to keep effective supervision on the distribution.

As the most similar negative pair and the positive pair influence the decision boundary the most, our contrastive loss updates the margin $m$ with the difference between the similarities of them.
The margin assures that the magnitudes of the exponential of the numerator and the denominator in softmax are close, and keeps the loss value steady. To solve the noises brought by the extreme data, we employ the Exponential Moving Average (EMA)~\cite{huang_curricularface_2020}. Specifically, let $m^{(k)}_C$ be the average of the margin of the $k$-th batch with $m^{0}_C=0$, and $\alpha$ be the momentum parameter which is empirically set to 0.99. For a pair $(\boldsymbol{h}_i, \boldsymbol{h}_j)$ where $i<j$, $m^{(k)}_C$ is updated as:
\vspace{-.05in}
\begin{small}
    \begin{align}
        m^{(k)}_C &= \alpha m^{(k)} +(1-\alpha) m^{(k-1)}_C,
        \label{eq:ema}\\
        m^{(k)} &= \frac{1}{N} \sum^N_{i=1} \left(\mathrm{sim}(\boldsymbol{h}_i, \boldsymbol{h}_{N+i}) - Maxneg_i\right), 
        \label{eq:m}\\
        Maxneg_i &= \max(\mathrm{sim}(\boldsymbol{h}_i, \boldsymbol{h}_j)), 
                j \in [1, 2N], j \neq N+i.
        \label{eq:neg}
    \end{align}
\end{small}
where $N$ is the number of samples.

Taking $m$ as the difference between angles like ArcFace~\cite{deng_arcface_2019} is also a candidate approach. However, it changes the angle of the vector pairs directly, which need to include the triangle function and increases the complexity of the derivation. This results in nan value when being used as the contrastive loss.
To sum up, our adaptive margin-based contrastive loss can be formulated as
\begin{small}
    \begin{equation}
        \mathcal{L}_{C} = - \log 
        \frac{e^{s (\mathrm{sim}(\boldsymbol{h}_i, \boldsymbol{h}_{N+i})-m_C)}}
            {e^{s (\mathrm{sim}(\boldsymbol{h}_i, \boldsymbol{h}_{N+i})-m_C)} + \sum^{n}_{j=1,j\neq i, j \neq N+i}e^{s\cdot\mathrm{sim}(\boldsymbol{h}_i, \boldsymbol{h}_j)}}.
            \label{eq:cosface}
    \end{equation}
\end{small}


\textbf{Supervised Contrastive Mask.}
When we apply the contrastive regularization in the framework, the incompatibility between the classification methods and the contrastive learning becomes problematic.
The naive format of contrastive learning loss in Eq~\ref{eq:contrastive} considers all feature pairs $(\boldsymbol{h}_{i}, \boldsymbol{h}_{j})$ where $i<j, j \neq N + i$ as negative, and splits them up in the representation space. This would conflict with the fact that some samples are from the same class in FR, i.e. ${y}_i={y}_j$, and thus their features should be similar.
When cooperating with the classification loss, the two methods could disturb each other in the interpretation of the supervision signals.

To avoid the above conflict between the contrastive regularization and the classification loss, we ignore the relationship between images from the same class.
Specifically,
with the help of labels in the training process, we create a supervised contrastive mask (SCM) to exclude the distraction of these samples by setting their similarity score $\mathrm{sim}(\boldsymbol{h}_i, \boldsymbol{h}_j)$ to $0$, where $i<j,j \neq N+i,$ and ${y}_i = {y}_j$.
Thus, the classification method takes advantage of the label, while the contrastive learning regularizes the feature distribution separately with identity-free signals.

\begin{figure}
    \centering
    \includegraphics[scale=0.55]{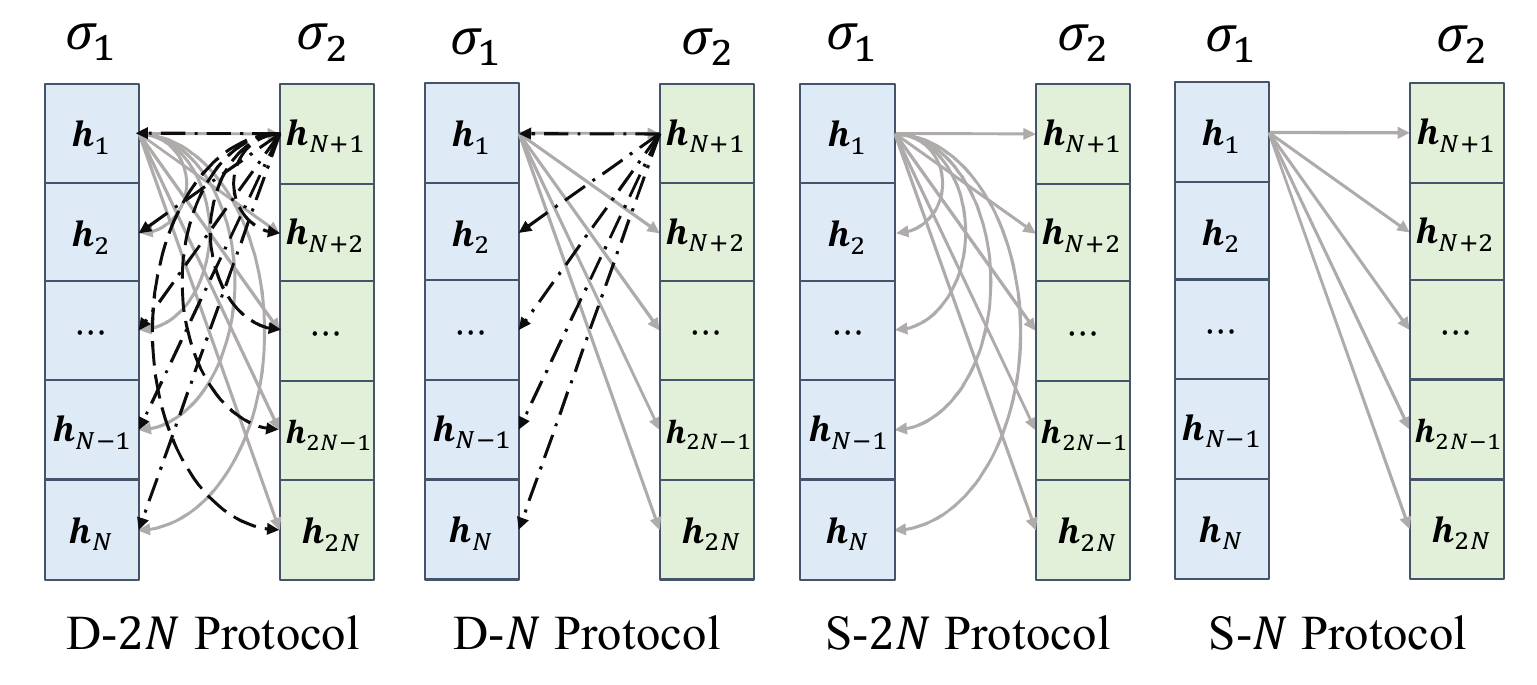}
    
    \caption{Four types of pair-coupling protocols. Every combination of two augmentation channels $\sigma_1$ and $ \sigma_2$ represents the feature combinations of the images in a mini-batch. When there are more than one augmentation channel combinations, the feature combination is repeated.
    }
    \label{fig:protocol}
    \vspace{-.25in}
\end{figure}

\subsection{Pair-Coupling Protocol for SRS Problem}
\vspace{-.05in}

By integrating our novel contrastive loss into training, the learned representation space could be constantly regularized. However, we discover a semantically repetitive signal problem (SRS) in this process. Some part of the contrastive loss is unintentionally repeated since some key negative pairs are doubly or quadruply emphasized. This results in a distorted distribution where the features encountering SRS are abnormally drawn and pulled.
To understand this problem, we investigate the pair-coupling protocols, i.e. how to compose the positive and negative pairs. Figure~\ref{fig:protocol} shows four different protocols.
Let $(\sigma_i \rightarrow \sigma_j)$ represents a pair where the first and the second features are from the $i$-th and the $j$-th mask channels respectively, and $i,j\in\{0,1\}$.
A pair-coupling protocol is defined by the number of the mask channels of the two features in the ordered pairs. Taking the second feature as the subordinate of the first one, the number of the first feature mask channels controls the ways~(\emph{single} or \emph{double}) and the mask channel of the second one dominates the number of negative samples~($N$ or $2N$).

\begin{figure}
    \centering
    \includegraphics[scale=0.43]{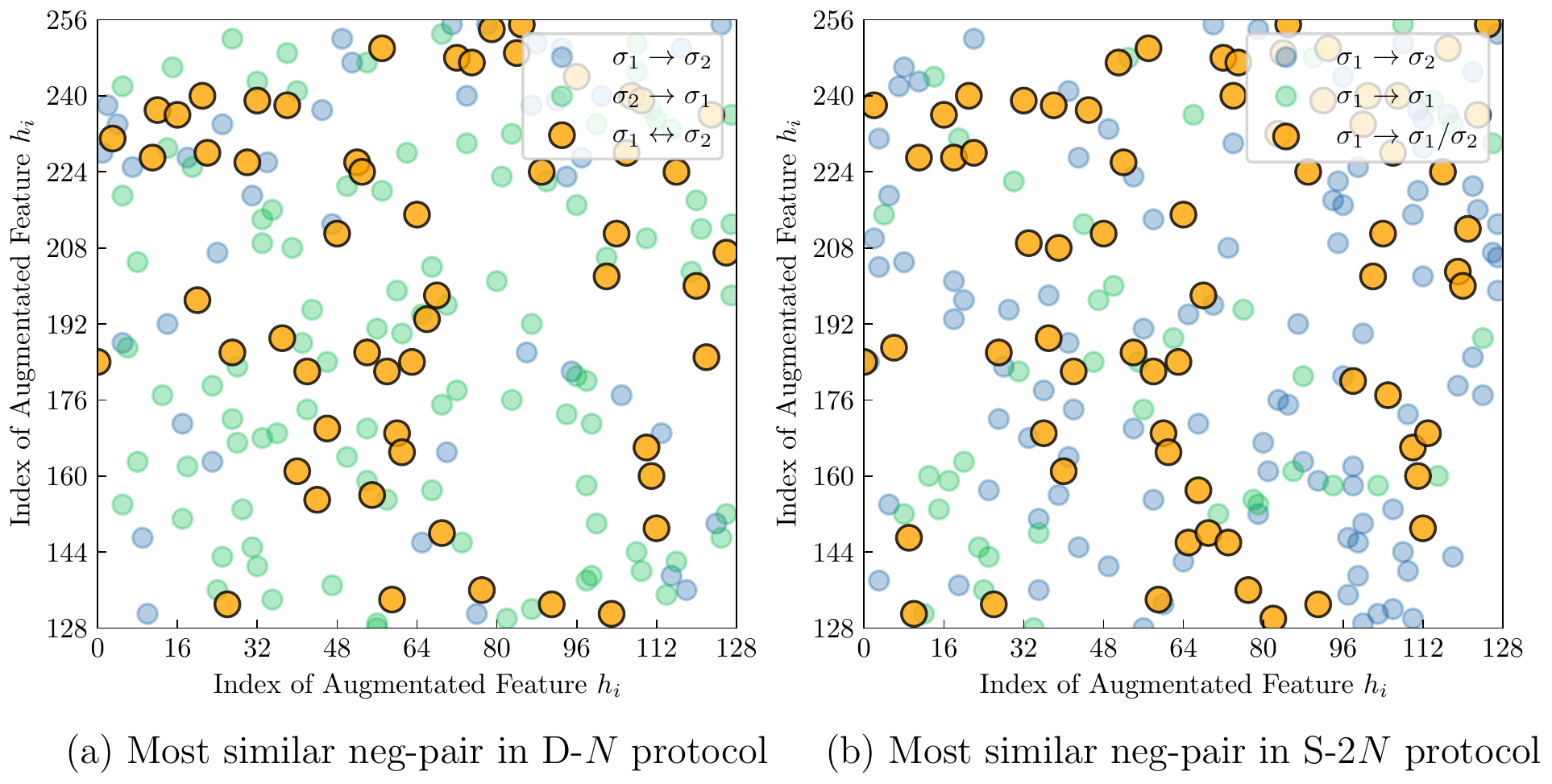}
    
    \caption{
    The coordinates of a point are the indexes of a feature and its most similar negative feature in a mini-batch.
    When two ordered pairs are mirrored, the points overlap and are painted yellow. The blossom yellow points in (a) and (b) demonstrate the symmetric problem in the ways and number of negative samples separately. The y-index of $(\sigma_1 \rightarrow \sigma_1)$ points are increased by 128.}
    \label{fig:index}
    \vspace{-.2in}
\end{figure}


\emph{Double-way $2N$ Protocol} is widely used in sample-guided contrastive learning, which takes the other $2N-1$ augmented features in the same mini-batch as the comparison pool of $\boldsymbol{h}$, and all $2N$ features from two augmentation channels are taken into the first position in a positive pair once. This protocol is completely symmetric, i.e. $(\sigma_{1}/\sigma_{2} \rightarrow \sigma_{1}/\sigma_{2})$. About $2\times N$ negative pairs in each of the two ways could quadruple the number of comparisons between every image pair.
When the most similar negative pairs of two given features are mirrored, the semantic effect of their relationship is doubled due to the repetitions. This property would result in a biased loss, which is undesired.

\begin{table*}
\centering
\setlength{\tabcolsep}{5mm}
\begin{small}
    \begin{tabular}{l|c|ccccc}
    \hline
    Methods (\%)     &Venue      & LFW   & AgeDB &CFP-FP & CALFW  & CPLFW 
    \\ 
    \hline
    CosFace         &CVPR~2018  & 99.81 & 98.11 & 98.12 & 95.76  & 92.28 
    \\
    ArcFace         &CVPR~2019  & {\textbf{99.83}} & 98.28 & 98.27 & 95.45  & 92.08 
    \\
    MV-Softmax      &AAAI~2020  & 99.80 & 97.95 & 98.28 & 96.10  & 92.83 
    \\
    CurricularFace  &CVPR~2020  & 99.80 & 98.32 & 98.37 & {\textbf{96.20}}  & 93.13 
    \\
    SCF-ArcFace     &CVPR~2021  & 99.82 & 98.30 & 98.40 & 96.12  & 93.16 
    \\
    MagFace         &CVPR~2021  & {\textbf{99.83}} & 98.17 & 98.46 & 96.15  & 92.87 
    \\
    AdaFace         &CVPR~2022  & 99.82 & 98.05 & 98.49 & 96.08	 & {\textbf{93.53}}	 
    \\
    \hline
    CoReFace      &Ours       & {\textbf{99.83}} & {\textbf{98.37}} & {\textbf{98.60}} & {\textbf{96.20}}  & 93.27    
    \\
    \hline
    \end{tabular}
\end{small}
\caption{Verification accuracy (\%) on LFW, AgeDB, CFP-FP, CALFW, and CPLFW. The {\textbf{Best}} results are emphasized in bold.}
\label{tab:performance}
\vspace{-.15in}
\end{table*}

\begin{table}
\centering
\begin{small}
    \setlength{\tabcolsep}{1.1mm}
    
    \begin{tabular}{lllllll}
    \hline
    \multicolumn{1}{c}{Methods(\%)} & \multicolumn{3}{c}{IJB-B(TAR@FAR)} & \multicolumn{3}{c}{IJB-C(TAR@FAR)} \\ \cline{2-7}
                                        & \multicolumn{1}{c}{1e-6}  & \multicolumn{1}{c}{1e-5}  & \multicolumn{1}{c}{1e-4}  & \multicolumn{1}{c}{1e-6}  & \multicolumn{1}{c}{1e-5}  & \multicolumn{1}{c}{1e-4}\\
    \hline
    Softmax                             & 46.73 & 75.17 & 90.06 & 64.07 & 83.68 & 92.40 \\
    SphereFace~\cite{liu_sphereface_2017}& 39.40 & 73.58 & 89.19 & 68.86 & 83.33 & 91.77 \\
    CosFace~\cite{wang_cosface_2018}    & 40.41 & 89.25 & 94.01 & 87.96 & 92.68 & 95.56 \\
    ArcFace~\cite{deng_arcface_2019}    & 38.68 & 88.50 & 94.09 & 85.65 & 92.69 & 95.74\\
    SCF-ArcFace~\cite{scf}     &\multicolumn{1}{c}{-}&\textbf{90.68} & 94.74 &\multicolumn{1}{c}{-}&\multicolumn{1}{c}{94.04}& 96.09   \\
    Magface~\cite{meng_magface_2021}         & 42.32 & 90.36 & 94.51 & \textbf{90.24} & 94.08 & 95.97  \\
    \hline
    CoReFace      & \textbf{47.02} & 91.33 &\textbf{95.09} & 89.34 & \textbf{94.73} & \textbf{96.43} \\
    \hline
    \end{tabular}
\end{small}
\caption{1:1 verification on IJB-B and IJB-C.
}
\label{tab:IJB}
\vspace{-.25in}
\end{table}

Figure~\ref{fig:index} shows the repetitions of the key negative pairs from a well-trained classification model. The coordinates of a point represent the indexes of a feature and its most similar negative counterpart in a batch. The points are painted in blue or green depending on the feature channels of a pair. When two points overlap, the position is painted yellow. In Figure~\ref{fig:index}(a), many pairs composed by features from different channels, $(\sigma_1 \rightarrow \sigma_2)$ and $(\sigma_2 \rightarrow \sigma_1)$, are mirrored. As they share the same key negative pairs, their contributions are almost the same. Contrastive loss function would take them to produce a partly doubled loss value. When this accompanies the whole training process, it becomes an unintentional hard example mining strategy and inappropriately guides the back-propagation.
The same problem exists in the choice of the comparison pool $(\sigma_1 \rightarrow \sigma_1)$ and $(\sigma_1 \rightarrow \sigma_2)$ as shown in Figure~\ref{fig:index}(b).


To solve this problem, we cut down the symmetry in the pair-coupling process by proposing a \emph{Single-way $N$ Protocol}. Specifically, we only calculate the similarity of $(\sigma_1 \rightarrow \sigma_2)$ in a batch and ignore the other three compositions. In this way, no extra repeated loss would be calculated. This seems contradictory with the common contrastive learning setting that needs more negative samples for comparison~\cite{chen_simple_2020,moco}. However, these methods are usually supported by complex data augmentations and a large comparison pool. The stochasticity and the rich candidates provide more possibilities for a given feature. While in FR, the data augmentation is destructive and abandoned, and a huge batch (of size 8,192) is generally not applicable~\cite{chen_simple_2020}.

After applying the supervised contrastive mask and the single-way $N$ protocol, we update $Maxneg_i$ and our contrastive loss function as

\begin{scriptsize}
    \begin{align}
        &Maxneg_i = \max(\mathrm{sim}(\boldsymbol{h}_i, \boldsymbol{h}_j)), j \in [N\!+\!1, 2N], {y}_i \neq {y}_{j},\\
        &\mathcal{L}_{CoRe} = -\log 
        \frac{e^{s         (\mathrm{sim}(\boldsymbol{h}_i,\boldsymbol{h}_{N+i}) - m_C)
        }}
        {e^{s
        (\mathrm{sim}(\boldsymbol{h}_i,\boldsymbol{h}_{N+i})
        - m_C)
        } + 
            \underset{
                j = N+1, {y}_i \neq {y}_j
            }{\overset{2N}\sum}
            e^{s~\mathrm{sim}(\boldsymbol{h}_i,\boldsymbol{h}_j)}},\\
        &\mathcal{L} = \frac{1}{2}\left(\mathcal{L}_{Cla}(\boldsymbol{h}_i)+\mathcal{L}_{Cla}(\boldsymbol{h}_{N+i})\right) + \lambda\mathcal{L}_{CoRe}(\boldsymbol{h}_i, \boldsymbol{h}_{N+i}).
        \label{eq:CoReFace}
    \end{align}
\end{scriptsize}

\begin{table}
\centering
    \begin{small}
    \setlength{\tabcolsep}{6mm}
    \begin{tabular}{l|cc}
    \hline
    Methods (\%)    & Id      & Ver     \\ 
    \hline
    CosFace~\cite{wang_cosface_2018}        & 97.91   & 97.91   \\
    ArcFace~\cite{deng_arcface_2019}        & 98.35   & 98.48   \\
    MV-Softmax~\cite{wang_mis-classified_2020}     & 97.76  & 97.80   \\
    CurricularFace~\cite{huang_curricularface_2020} & \textbf{98.71}   & 98.64   \\ 
    BroadFace~\cite{kim_broadface_2020}      & 98.70   & 98.95    \\
    CircleLoss~\cite{circle_loss}      & 98.50   & 98.73    \\
    \hline
    CoReFace     & 98.69   & \textbf{99.06} \\ 
    \hline
    \end{tabular}
    \end{small}

\caption{Face identification and verification on MegaFace Challenge using FaceScrub as the probe set. Id refers to the rank-1 face identification accuracy with 1M distractors, and Ver refers to the face verification \textbf{TAR (@FAR=$10^{-6}$)}.}
\label{tab:megaface}
\end{table}

\section{Experiments}
\subsection{Datasets and Implementation Details}
\textbf{Datasets.} We use MS1MV2~\cite{deng_arcface_2019} for model training. MS1MV2 dataset contains about 5.8M face images of 85K individuals. We extensively evaluate our approach on eight benchmarks, including LFW~\cite{huang_lfw_2007}, AgeDB~\cite{agedb}, CFP-FP~\cite{cfp-fp}, CPLFW~\cite{cplfw}, CALFW~\cite{calfw}, IJB-B~\cite{ijbb}, IJB-C~\cite{ijbc}, and MegaFace~\cite{megaface}.


\textbf{Training Settings.} We follow the settings commonly used in recent works~\cite{wang_mis-classified_2020,kim_adaface_2022,kim_broadface_2020,huang_curricularface_2020, meng_magface_2021} to ensure the fairness of comparison. The face images are cropped and resized to $112 \times 112$ with five landmarks~\cite{deng_arcface_2019}. 
We employ ResNet100~\cite{resnet} as the backbone model. ArcFace is employed as the classification loss. Our framework is implemented in Pytorch~\cite{pytorch}. We train the models on 4 NVIDIA A100 GPUs with the batch size of 512. All models are trained using SGD algorithm with an initial learning rate of $0.1$. We set the momentum to 0.9 and the weight decay to $5\times 10^{-4}$. We divide the learning rate by $10$ at the 8th, the 14th, and the 20th epochs, and stop the training after 24 epochs.
We set the scale parameter $s$ to 64 for both the classification loss and our loss, and set $\lambda$ to 0.05.
For fair comparison of evaluation results, all methods without specifications are implemented with ResNet100 and MS1MV2.



%
%

\begin{table}
    \centering
    \begin{small}

    \begin{tabular}{c|l|c}
    \hline
    Setting Groups & \multicolumn{1}{c|}{Methods}   & Average  \\ 
    \hline
     \multirow{2}{*}{\begin{tabular}[c]{@{}c@{}}Single Supervision\end{tabular}}
     & Classification-only      & 93.60  \\
     & Triplet-only             & 91.03 \\
    \hline
    \multirow{3}{*}{\begin{tabular}[c]{@{}c@{}}Contrastive \\Only\end{tabular}}
     & NT-Xent        & 63.61 \\
     & SupCon         & 67.87  \\
     & CoReFace       & 86.68   \\
    \hline
     \multirow{2}{*}{\begin{tabular}[c]{@{}c@{}}Data \\Augmentation\end{tabular}}
    & NT-Xent                & 92.78 \\
    & SupCon                & 91.49 \\
    \hline
    \multirow{3}{*}{\begin{tabular}[c]{@{}c@{}}Feature \\Augmentation\end{tabular}}
     & NT-Xent                  & 93.60 \\
     & SupCon                   & 93.60 \\
     & CoReFace          & \textbf{93.66} \\
    \hline
    \end{tabular}
    \end{small}
    \caption{Average verification performance (\%) of different methods.
    All experiments are based on a pretrained ResNet50 ArcFace model with 90.45\% average performance.
    To avoid the influence of hyper-parameter, $\lambda=1$ is set for all experiments.}
    \label{tab:methods}
\end{table}

\subsection{Experiment Results}

\textbf{Results on LFW, CFP-FP, AgeDB, CALFW and CPLFW.} Table~\ref{tab:performance} compares our CoReFace with other recent approaches on diverse benchmarks, including LFW for unconstrained face verification, AgeDB and CALFW of various ages, CFP-FP and CPLFW with large pose variations.
In our approach, ArcFace is employed as the classification loss.
Compared to the original ArcFace, our CoReFace outperforms it on four out of the five datasets with remarkable margins and achieves the same performance on the last one. 
This is because that CoReFace incorporates contrastive regularization in representation learning, which can successfully address the aforementioned inconsistency problem between the identify-based training and the sample-based evaluation which is ignored in the existing approaches.
Among all approaches, AdaFace takes the image quality into consideration during training.
This could explain its superior performance on CPLFW where different poses may cause occlusions on faces and results in lower accuracy.
Our method strikes the highest accuracies on the other four datasets. Especially when our CoReFace shares the top performance with ArcFace and CurricularFace on LFW and CALFW, we signicantly outperform them on the other datasets.



\textbf{Results on IJB-B and IJB-C.}
The IJB-B dataset contains $1,845$ subjects with 21.8K still images and 55K frames from $7,011$ videos. The IJB-C dataset expands IJB-B, and contains about $3,500$ identities with a total of 31.3K images and 117.5 unconstrained video frames. In the 1:1 verification task, there are about 10K positive matches and 8M negative matches in IJB-B, and 19K positive matches and 15M negative matches in IJB-C.
Table~\ref{tab:IJB} exhibits the performances of different methods for 1:1 verification on IJB-B and IJB-C. Our method achieves the highest \textbf{TAR}s for two out of three 
different \textbf{FAR}s on these two datasets separately. As IJB-B has fewer matches, it becomes the most challenging situation when $\textbf{FAR}=10^{-6}$ and only about 8 negative matches are allowed. Compared with other methods whose \textbf{TAR}s are lower than Softmax, our model demonstrates more competitive under such an extreme situation. When there is a higher \textbf{FAR} bound ($10^{-4}$) or the evaluation dataset is larger, CoReFace still outperforms the competitors.

\begin{figure}
    \centering
    \includegraphics[scale=0.38]{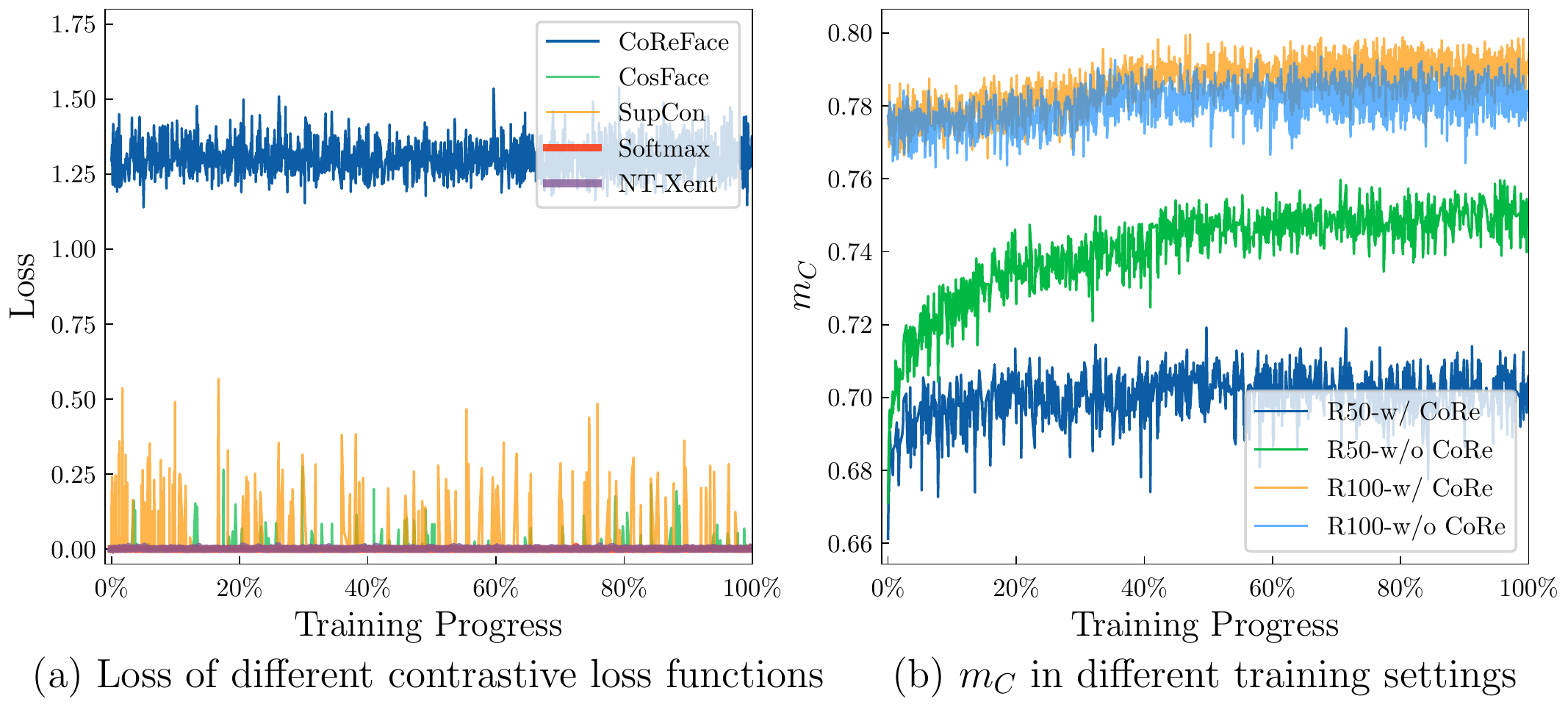}
    
    \caption{(a) The loss variation of different contrastive methods in joint training with R100. (b) The $m_C$ value variation caused by CoReFace on different backbone models. Some methods keep their loss values nearly 0 and fail to supervise in training.}
    \label{fig:loss_m}
\end{figure}

\begin{table}
    \centering
    \setlength{\tabcolsep}{1.1mm}
    \begin{small}
    \begin{tabular}{c|c|c|c|c|c|c}
    \hline
    \multirow{2}{*}{\begin{tabular}[c]{@{}c@{}}SCM\end{tabular}}
    & \multirow{2}{*}{\begin{tabular}[c]{@{}c@{}}Original\end{tabular}}   
    & \multirow{2}{*}{\begin{tabular}[c]{@{}c@{}}w/o $L_C$\end{tabular}}
    & \multicolumn{4}{c}{w/ $L_C$}  \\
    \cline{4-7}
      &       &         & D-2$N$   & D-$N$   &S-2$N$   &S-$N$    \\
    \hline
    \ding{55}    &93.28  &93.59  &93.66  &93.68  &93.67  &\textbf{93.74} \\
    \cline{2-7}
    \ding{51}    &-      &-      &93.65  &93.70  &93.69  &\textbf{93.75} \\
    \hline
    \end{tabular}
    \end{small}
    \caption{Ablation of different pair-coupling protocols and supervised contrastive mask. All experiments except \textit{Original} are implemented with the proposed framework. \textit{S} and \textit{D} mean \textit{single-way} and \textit{double-way} respectively. $N$ and 2$N$ represent the number of candidates for a given sample.}
    \label{tab:abla_method}
    \vspace{-.15in}
\end{table}

\textbf{Results on MegaFace.} Finally, we demonstrate the efficacy of our method on the MegaFace Challenge. The gallery set of MegaFace contains 1M images of 690K subjects, and the probe set is FaceScrub, which contains 100K photos of 530 unique individuals.
We follow~\cite{deng_arcface_2019} to remove the face images with wrong labels and evaluate our method on the refined dataset.
Table~\ref{tab:megaface} shows the performance of different methods.
For the identification task, CoReFace achieves competitive performance which is only 0.02\% lower compared to the highest one CurricularFace~\cite{huang_curricularface_2020}.
For the verification task, CoReFace outperforms all the other approaches with a clear margin.
The BroadFace~\cite{kim_broadface_2020} also shows competitive performance by building a dynamic queue to gain extra training on the classification layer. Without complex structure reformation, CoReFace adds an image-image regularization to improve the feature distribution and boost the performance of open-set face recognition.


\subsection{Ablation Study}

As LFW is an almost saturated dataset (the accuracy is about 99.8\% with ResNet100), we report the performances on AgeDB, CFP-FP, CALFW, CPLFW, and their average in our ablation study.

\textbf{Effects of our CoReFace Loss Function.}
We show the effectiveness of our contrastive loss by comparing it with other alternatives with different settings in Table~\ref{tab:methods}. The \emph{Contrastive Only} group is apparently inferior to the classification methods, which demonstrates the necessity to take the classification loss as the fundamental in FR.
Compared with the classification-only method, the performance degradation of NT-Xent and SupCon in \emph{Data Augmentation} group verifies the semantic damage caused by the widely-used image augmentations.
The \emph{Feature Augmentation} group follows our framework and CoReFace shows an outstanding outcome.

Figure~\ref{fig:loss_m} further visualizes the contrastive loss values and the adaptive margin $m_C$ during joint training. With the adaptive margin, our method produces stable and reasonable loss values.
NT-Xent and SupCon in \emph{Feature Augmentation} group perform similarly compared with the Classification-only approach. Figure~\ref{fig:loss_m} illustrates how they fail to supervise steadily. 
The change of $m_C$ with different backbones confirms the adaptation of our method which saves tedious hyper-parameter tuning for different model scales. In our CoReFace, $m_C$ obviously keeps growing, and surpasses the one without $m_C$ in R50 where it is only statistically calculated in training. This verifies that our contrastive loss can effectively enlarge the difference between the similarities of the positive pairs and the negative pairs.



\begin{table}
    \centering
    \setlength{\tabcolsep}{1.1mm}
    \begin{small}
    
    \begin{tabular}{c|c|c|c|c}
    \hline
      Time (batch/s)  & Original & CoReFace & Contrastive &  Triplet\\
    \hline
    \multicolumn{1}{c|}{R50}  & 0.2807 & 0.2811 & 0.5052 & 0.7104  \\ 
    \multicolumn{1}{c|}{R100} & 0.4874 & 0.4890 & 0.8465 & - \\ 
    \hline
    \end{tabular}
    
    \end{small}
    \caption{Average process time for a batch in each method on one NVIDIA A100 GPU.
    We take triplets as samples for triplet loss. R100 with triplet loss needs more than 40GB video memory to train such a batch and fails in the training.
    }
    \label{tab:speed_batch}
\end{table}

\textbf{Effects of other components.}
To verify the effect of our framework, we generally experiment three different settings, namely \textit{Original, w/o $L_C$}, and \textit{w/ $L_C$} in Table~\ref{tab:abla_method}. \textit{Original} means the traditional classification framework and the latter two take our framework.
Among the four pair-coupling protocols, the D-$2N$ protocol, the D-$N$ protocol, and the S-$2N$ protocol all contain some repetitive key negative pairs. Table~\ref{tab:abla_method} shows that the average performance of D-$2N$ ranks the lowest as it contains more repetitive pairs while the single-way $N$ protocol outperforms the others. These results verify our assumption that the symmetry in pair coupling interferes the performance. 
We also implement a series of experiments without the supervised contrastive mask under each pair-coupling protocols, which show that the masked version outperforms the others most time. Though the sampling in training is stochastic and results in a few conflicts, the mask still expels the supervision conflict between two losses.

\textbf{Efficiency of different frameworks.}
Table~\ref{tab:speed_batch} compares the training speed of different frameworks including \textbf{Original} classification framework, our feature-augmentation-based \textbf{CoReFace} framework, data-augmentation-based \textbf{Contrastive} framework, and \textbf{Triplet} frameworks.
As can be seen in Table~\ref{tab:speed_batch}, after integrating the contrastive regularization into training, our framework only takes negligible extra time, i.e. 1.4\textperthousand~for R50 and 3.3\textperthousand~for R100 compared to the original classification method. Meanwhile, the common contrastive framework nearly doubles the processing time.
As for Triplet, it fails to be applied on a 40G GPU with R100 and consumes a lot more time on CASIA-WebFace.

\begin{figure}
    \centering
    \includegraphics[scale=0.38]{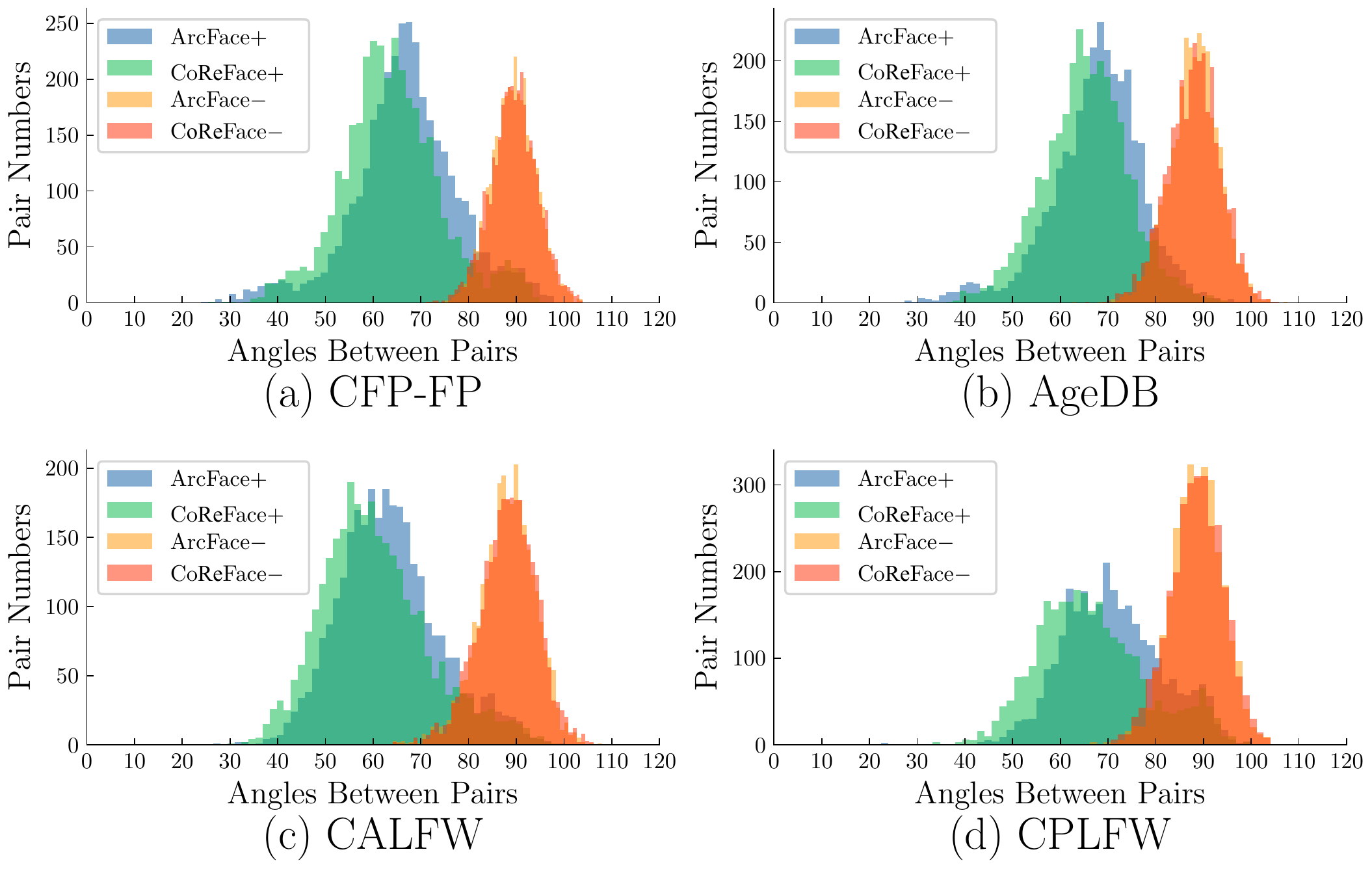}
        \caption{The angle distributions of ArcFace and CoReFace on four datasets. $+$ and $-$ denote the positive pairs and the negative pairs respectively.}
    \label{fig:analysis}
    \vspace{-.2in}
\end{figure}

\textbf{Effects on the feature distribution.}
We visualize the similarity between the positive and the negative pairs on the evaluation datasets in
Figure~\ref{fig:analysis}. The angles of positive pairs in CoReFace is closer to 0 compared with ArcFace. For different datasets containing age variations and pose variations, our approach keeps an obvious margin.
This demonstrates the effectiveness of our CoReFace for the distribution regularization by considering the image-image relationship.

\section{Conclusion}
We have presented our CoReFace to regularize the feature distribution with the image-image relationship, which makes the training consistent with the evaluation in open-set face recognition. To this end, the sample-guided contrastive learning is integrated in our framework.
For positive pair composition in contrastive learning, we augment the embeddings instead of images
and avoid the degradation caused by the widely-used data augmentations.
By incorporating an adaptive margin and a supervised contrastive mask, our contrastive loss is able to generate steady loss values and avoid the collision with the classification supervision signals.
Finally, the new pair-coupling protocol alleviates the similarity problem caused by the symmetry of pairs.
Extensive experiments on the popular FR benchmarks and ablations demonstrate the effectiveness and efficiency of our proposed approach and the great potential of contrastive learning for regularization in face recognition.
With the concise framework, our approach can be easily applied to the existing FR methods. 

{\small
\bibliographystyle{ieee_fullname}
\bibliography{egbib}
}

\end{document}


\title{CoReFace: Sample-Guided Contrastive Regularization for Deep Face Recognition}

\author{First Author\\
Institution1\\
Institution1 address\\
{\tt\small firstauthor@i1.org}
\and
Second Author\\
Institution2\\
First line of institution2 address\\
{\tt\small secondauthor@i2.org}
}

\maketitle
\ificcvfinal\thispagestyle{empty}\fi

\section*{A. Data Augmentation on Face}
\begin{figure}[ht]
    \centering
    \includegraphics[scale=0.4]{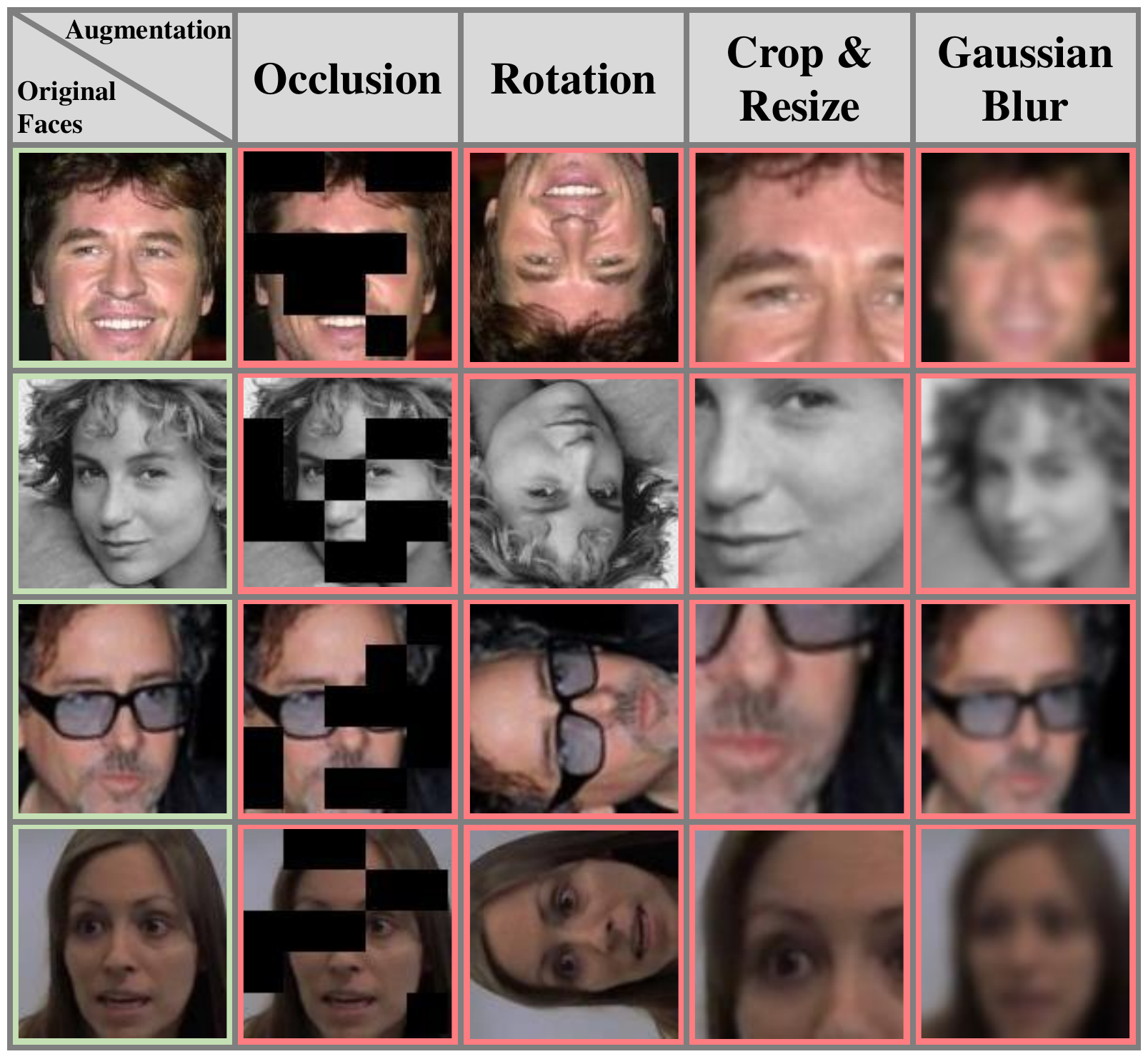}
    \caption{Image augmentation on Face.}
    \label{fig:head}
\end{figure}
\cref{fig:head} shows some examples of the distortions caused by data augmentation on face images. The commonly used occlusion, rotation, cropping with resizing, and Gaussian blur degrade the recognizability of faces. The combination of these augmentations, which is a key component of sample-guided contrastive learning, could make them more confusing. Furthermore, some face images~(both in the training dataset and evaluation dataset) could be blurry, abnormally exposed, twisted posed or with facial decorations. This could decrease the effective identification area of a face image. 

\section*{B. Ablation on CASIA-WebFace}

\begin{table}[h]
\centering
\begin{small}
\setlength{\tabcolsep}{0.5mm}{
\begin{tabular}{c|c|cccc|c}

\hline
$\mathrm{dropout}_1$ & $\mathrm{dropout}_2$ & AgeDB  & CFP-FP & CALFW  & CPLFW  & Average  \\ \hline
0.1                  & 0.9  & 95.02 & 95.50 & 94.07 & 90.60 & 93.80 \\
\hline
\multirow{3}{*}{0.2} & 0.7  & 95.00 & 95.66 & 94.12 & 90.45 & 93.81 \\
                     & 0.8  & 94.87 & 95.54 & 94.22 & 90.73 & 93.84 \\
                     & 0.9  & 95.02 & 95.73 & 94.40 & 90.55 & \textbf{93.92} \\
\hline
0.3                 & 0.9  & 94.83 & 95.51 & 94.17 & 90.45 & 93.74 \\
\hline
0.4                 & 0.9  & 95.20 & 95.66 & 93.97 & 90.27 & 93.77 \\
\hline
0.5                 & 0.9  & 95.08 & 95.69 & 93.92 & 90.42 & 93.78 \\
\hline

\end{tabular}
}
\end{small}
\caption{Verification performances (\%) with different combinations of dropout probabilities in two stages.}
\label{tab:dropout}
\end{table}

\begin{table}[h]
\centering
\begin{small}

\begin{tabular}{c|cccc|c}
\hline
$\lambda$  & AgeDB  & CFP-FP  & CALFW   & CPLFW   & Average  \\ 
\hline
0.1    & 95.03  & 95.47   & 94.02   & 90.37   & 93.72  \\
0.3    & 95.00  & 95.69   & 94.18   & 90.48   & 93.84  \\
0.5    & 95.02  & 95.73   & 94.40   & 90.55   & \textbf{93.92}  \\
0.7    & 94.93  & 95.60   & 94.27   & 90.65   & 93.86  \\
1.0    & 95.05  & 95.56   & 94.12   & 90.60   & 93.83  \\
\hline
\end{tabular}

\end{small}
\caption{Verification performances (\%) with different $\lambda$.}
\label{tab:lambda}
\end{table}

For training with R50 and CASIA-WebFace, we carry out ablation study on the hyper parameters
as shown in \cref{tab:dropout} and \cref{tab:lambda}. The best dropout probability and $\lambda$ vary from the setting of R100 with MS1MV2. A larger dropout probability gap between two stages benefits the training more. The extreme value 0.9 even performs the best.

The $\lambda$ here is 10 times the one for R100. It could be explained that the larger network enjoys better feature representation and commits less mistake, while our ContraFace with the adaptive margin keeps generating almost the same loss value.

\begin{table*}[h]
\centering

\begin{tabular}{c|c|c|c|c|c|c}
\hline
\multirow{2}{*}{\begin{tabular}[c]{@{}c@{}}Time (ms/sample)\end{tabular}}
&\multirow{2}{*}{\begin{tabular}[c]{@{}c@{}}Original\end{tabular}}
&\multirow{2}{*}{\begin{tabular}[c]{@{}c@{}}ContraFace\end{tabular}}
&\multirow{2}{*}{\begin{tabular}[c]{@{}c@{}}Contrastive\end{tabular}}
&\multicolumn{3}{c}{Triplet}\\
\cline{5-7}
    &   &   &   & Original & Shrunken & Random
\\
\hline
\multicolumn{1}{c|}{R50 + CASIA-WebFace}  & 2.193 & 2.196 & 3.947 & 1.850 & 2.404 & 2.246  \\
\hline
\multicolumn{1}{c|}{R100 + MS1MV2} & 3.808 & 3.820 & 6.614 & -     & 16.70& 3.802 \\ 
\hline
\end{tabular}
\caption{Average process time for a sample in each method on one NVIDIA A100 GPU.}
\label{tab:speed-sample}
\end{table*}

\begin{table*}[h]
\centering

\begin{tabular}{c|c|c|c|c|c|c|c}
\hline
\multicolumn{2}{c|}{\multirow{2}{*}{}} &\multicolumn{3}{c|}{CALFW} & \multicolumn{3}{c}{CPLFW}\\
\cline{3-8}
\multicolumn{2}{c|}{} & ArcFace & ContraFace &Diff. &ArcFace &ContraFace &Diff.\\
\hline
\multirow{2}{*}{\begin{tabular}[c]{@{}c@{}}R50\end{tabular}}
&$+$  &1322.0 &1456.0 &$\uparrow$10.1\%   &961.4    &1172.4 &$\uparrow$21.9\%\\
&$-$  &99.4   &93.0   &$\downarrow$6.6\%   &73.7   &78.6   &$\uparrow$6.6\%\\
\hline
\multirow{2}{*}{\begin{tabular}[c]{@{}c@{}}R100\end{tabular}}
&$+$  &1769.1 &1744.2 &$\downarrow$1.4\%  &1417.7 &1380.5 &$\downarrow$0.4\%\\
&$-$  &27.0   &26.9   &$\downarrow$2.6\%   &13.4   &11.9   &$\downarrow$11.2\%\\
\hline
\end{tabular}
\caption{The similarity score of ArcFace and ContraFace on CALFW and CPLFW with R50 and R100. $+$ and $-$ denote the positive pairs and negative pairs respectively.}
\label{tab:similarity}
\end{table*}

\section*{C. Training Speed of a Sample}
The training speed is reported based on a batch with the size 128 in the main paper. \cref{tab:speed-sample} further reports the training time for a single sample with more settings on triplet loss. We call the aforementioned \emph{triplets as samples} Triplet-Original here. To make the samples in a batch generally equal, we shrink the batch size of triplet loss to 42 and it possesses a similar number of samples comparing with other methods, namely Triplet-Shrunken. Then, a simplified version is called Triplet-Random. The negative pairs are composed with samples extracted from a pool without the images from the same class. In Triplet-Original and Triplet-Shrunken, this is applied by a set subtraction with numpy, which costs most time. We then sample these negative samples randomly from the whole dataset with a Shrunken batch and name it Triplet-Random.

Again, our ContraFace shows a negligible difference comparing with the original classification training process. As for Triplet-Original, containing $3\times$ batch compared with the Original, it performs faster on a single sample. The Triplet-Shrunken releases triplet loss from video memory explosion, but it still suffers from the negative sampling. The Triplet-Random does not have the above problems. However, its function is similar with SupCon, which only contributes with the positive pair in the training and makes little improvement.
\section*{D. Similarity Score Changes}
After averaging the similarity score of two methods on two datasets with two backbones, we find that ContraFace changes the similarity distribution differently on R50 and R100. \cref{tab:similarity} shows the details. Comparing with ArcFace, the subtraction of the similarity score summation on positive pairs and negative pairs is raised by more than 15\% on CALFW and CPLFW with R50 as backbone. This shows the strong effect of redistribution with our method on a relative small model.
For R100, the model representation ability is already sound enough and our method decreases the similarity globally. However, it still imposes greater penalties on the negative pairs.

\section*{E. Details of Method Comparison}
Table~4 of method comparison in the main paper contains three SupCon outcomes. The two in \emph{Data Augmentation} Group are implemented with image augmentation. The original classification-based training process only contains \emph{Random Horizontal Flip}. The \emph{Data Augmentation} Group includes \emph{Random Resize\&Crop}, \emph{Random Horizontal Flip}, \emph{Color Distortion} and \emph{Gaussian blur}. The cropping scale is set to 0.2-1.0 for SupCon and 0.8-1.0 for SupCon-soft respectively. The brightness, contrast, and saturation are set to 0.4 for SupCon and 0.2 for SupCon-soft. We change these two augmentations and observe a 3\% improvement. Nevertheless, they are still obviously inferior to the classification-only training.

\section*{F. Visualization of Face Cases}
\begin{figure*}[ht]
    \centering
    \includegraphics[scale=0.4]{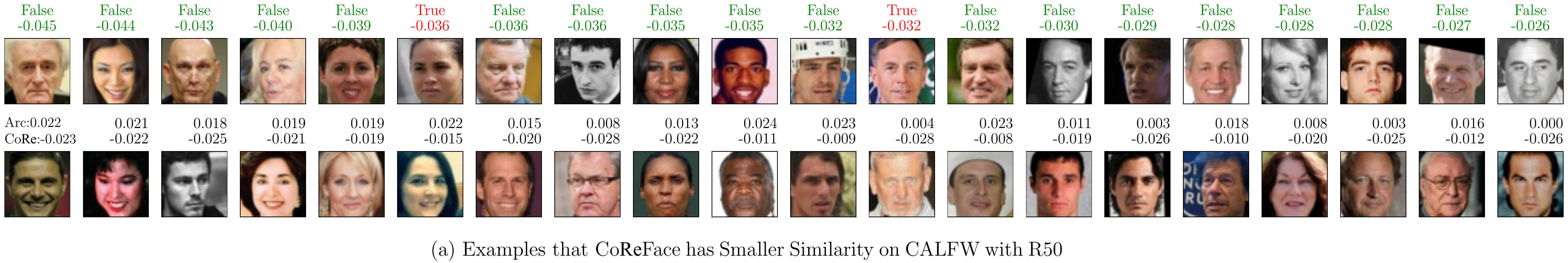}
    \includegraphics[scale=0.4]{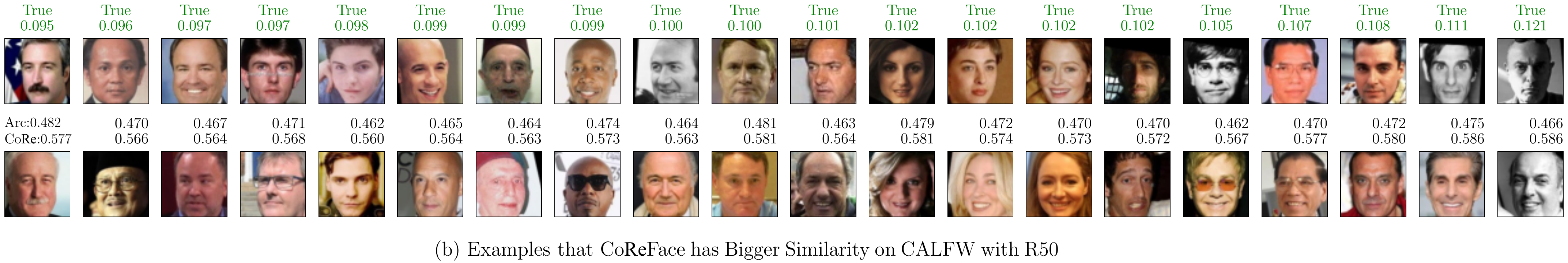}
    \includegraphics[scale=0.4]{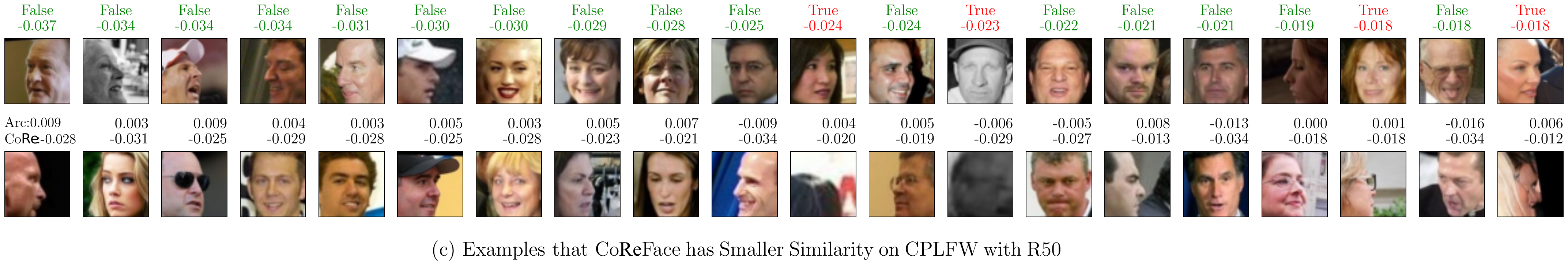}
    \includegraphics[scale=0.4]{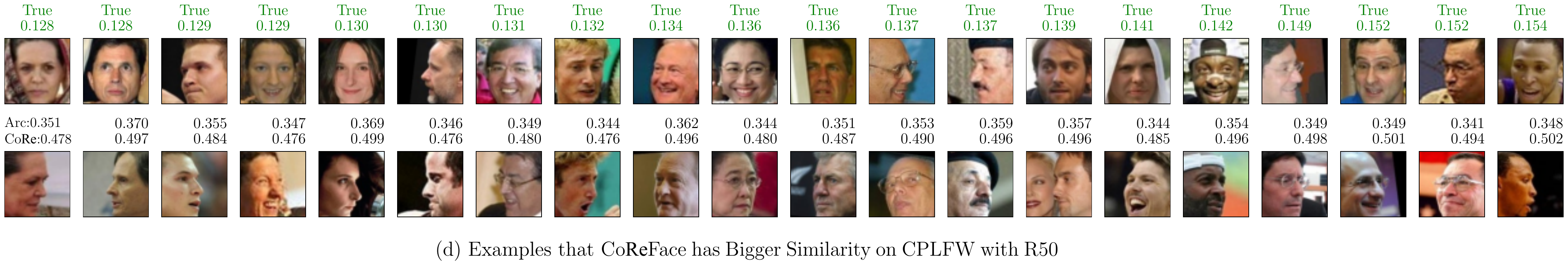}
    \caption{Examples of the difference between ArcFace and ContraFace with R50 as backbone. The label and the similarity subtraction are put on the top of a pair. The similarity scores of the two methods are put in the middle of a pair. If ContraFace has a proper change on the similarity based on the pair label, it will be painted green. Otherwise, we show it in red. Samples in (a) and (c) are from the intersection of 15\%$\sim$25\% of the similarity distributions of the two methods, while the ones in (b) and (d) are from the intersection of 75\%$\sim$85\% the similarity distributions of the two methods.}
    \label{fig:r50}
\end{figure*}

\begin{figure*}[ht]
    \centering
    \includegraphics[scale=0.4]{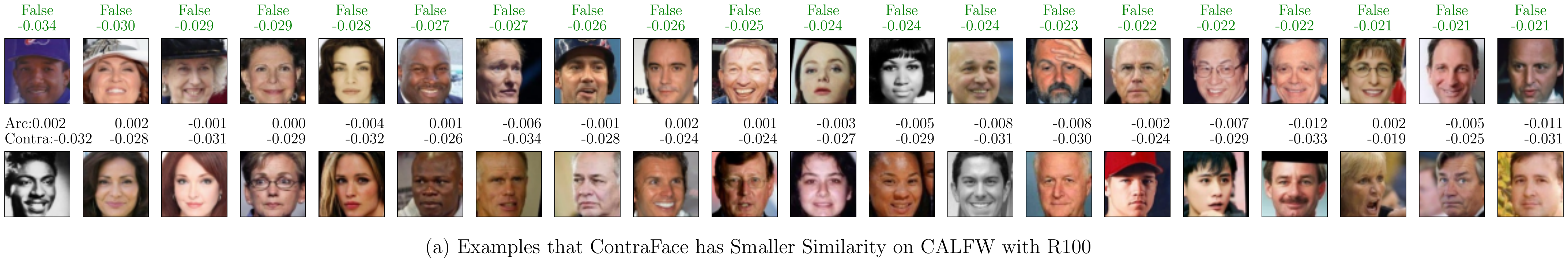}
    \includegraphics[scale=0.4]{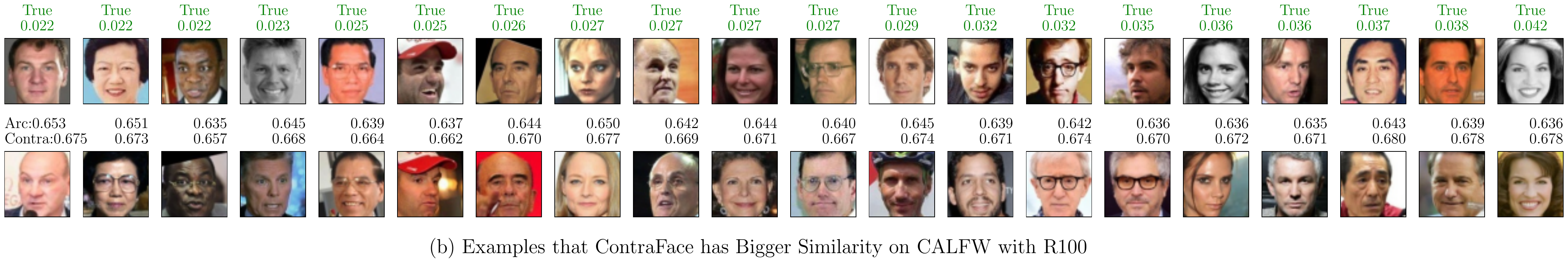}
    \includegraphics[scale=0.4]{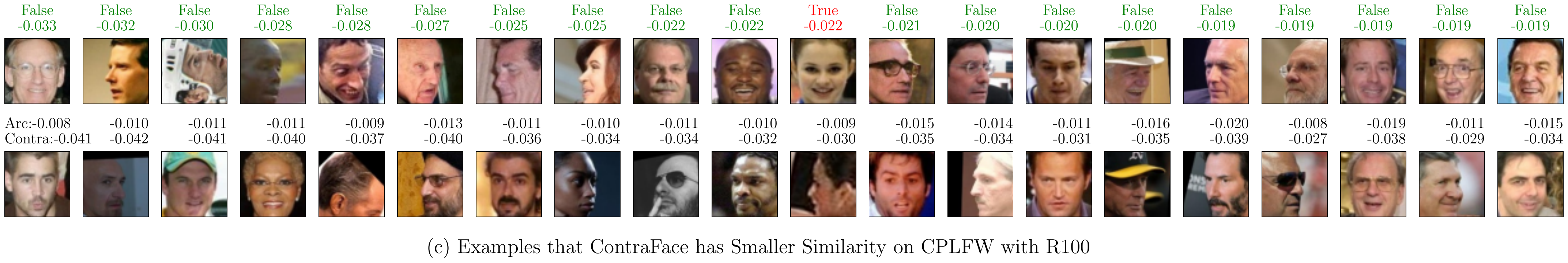}
    \includegraphics[scale=0.4]{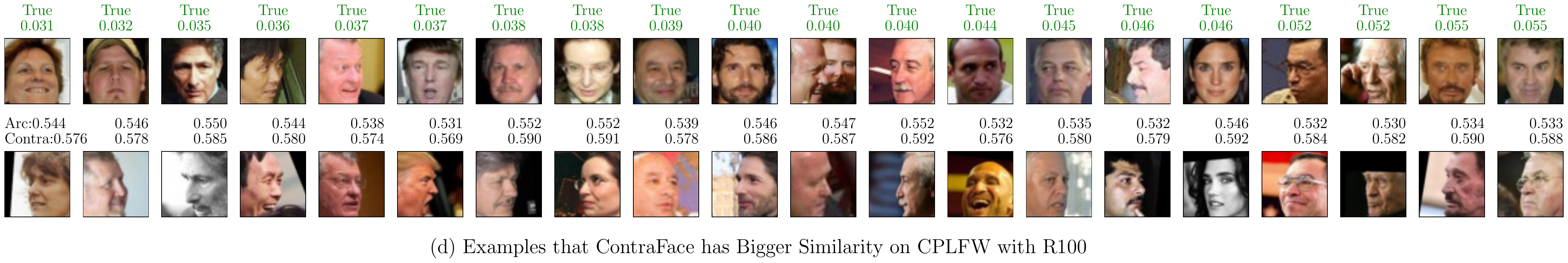}
    \caption{Examples of the difference between ArcFace and ContraFace with R100 as backbone. The label and the similarity subtraction are put on the top of a pair. The similarity scores of the two methods are put in the middle of a pair. If ContraFace has a proper change on the similarity based on the pair label, it will be painted green. Otherwise, we show it in red. Samples in (a) and (c) are from the intersection of 15\%$\sim$25\% of the similarity distributions of the two methods, while the ones in (b) and (d) are from the intersection of 75\%$\sim$85\% the similarity distributions of the two methods.}
    \label{fig:r100}
\end{figure*}

In \cref{fig:r50} and \cref{fig:r100}, we show the face image pairs and the similarity scores of them computed by ArcFace and ContraFace. Images are from the intersection of the same similarity distributions segment of the two methods. The 20 pairs with the largest absolute values of the subtraction are shown.
For the samples where ContraFace has smaller scores, both two methods give low accuracy. The main difference is that our method produces more and lower negative scores. It could be seen as expanding the boundaries of similarity. Generally, ContraFace shows its superiority on different poses and age gaps, especially with a relative small backbone.

